\documentclass[runningheads]{llncs}
\usepackage[T1]{fontenc}
\usepackage{amssymb}
\usepackage{graphicx,verbatim}
\usepackage{color}
\usepackage{hyperref}
\begin{document}
\title{MedDiff-FT: Data-Efficient Diffusion Model Fine-tuning with Structural Guidance for Controllable Medical Image Synthesis}
\titlerunning{MedDiff-FT}
% \title{MedDiff-FT: Fine-tuning diffusion model to generate controllable images for improving medial image segmentation}
%
% \begin{comment}  %% Removed for anonymized MICCAI 2025 submission

% \end{comment}

% \author{Anonymized Authors}  %% Added for anonymized MICCAI 2025 submission
% \authorrunning{Anonymized Author et al.}
\author{Jianhao Xie\inst{1} \and
Ziang Zhang\inst{1} \and
Zhenyu Weng\inst{2} \and
Yuesheng Zhu\inst{1} \and
Guibo Luo\thanks{Corresponding author: luogb@pku.edu.cn}\inst{1}}

\authorrunning{Jianhao, et al.}

\institute{Guangdong Provincial Key Laboratory of Ultra High Definition Immersive Media Technology, Peking University Shenzhen Graduate School\and
South China University of Technology, China}

\maketitle              % typeset the header of the contribution
\begin{abstract}
Recent advancements in deep learning for medical image segmentation are often limited by the scarcity of high-quality training data. While diffusion models provide a potential solution by generating synthetic images, their effectiveness in medical imaging remains constrained due to their reliance on large-scale medical datasets and the need for higher image quality. To address these challenges, we present \textbf{\emph{MedDiff-FT}}, a controllable medical image generation method that fine-tunes a diffusion foundation model to produce medical images with structural dependency and domain specificity in a data-efficient manner. During inference, a dynamic adaptive guiding mask enforces spatial constraints to ensure anatomically coherent synthesis, while a lightweight stochastic mask generator enhances diversity through hierarchical randomness injection. Additionally, an automated quality assessment protocol filters suboptimal outputs using feature-space metrics, followed by mask corrosion to refine fidelity. Evaluated on five medical segmentation datasets, \textbf{\emph{MedDiff-FT}}’s synthetic image-mask pairs improve SOTA method's segmentation performance by an average of 1$\%$ in Dice score. The framework effectively balances generation quality, diversity, and computational efficiency, offering a practical solution for medical data augmentation. The code is available at \href{https://github.com/JianhaoXie1/MedDiff-FT}{https://github.com/JianhaoXie1/MedDiff-FT}.
% Authors must provide keywords and are not allowed to remove this Keyword section.

\end{abstract}
\section{Introduction}
Medical image segmentation is a key step in medical image processing and analysis, which involves separating and extracting specific structures or regions (e.g., organs, tissues, lesions, etc.) from the background or other structures in a medical image. In recent years, the development of deep learning, especially convolutional neural networks \cite{cnn,resnet}, has greatly advanced medical image segmentation. Early approaches like U-Net \cite{unet} introduced an encoder-decoder architecture with cross-layer connections, achieving remarkable success in medical imaging. Subsequent advancements such as nnU-Net \cite{nnunet} further optimized pre-processing and network adaptation based on medical data characteristics. 
\begin{figure*}[htb]
\centering
\includegraphics[width=0.9\textwidth,height=6cm]{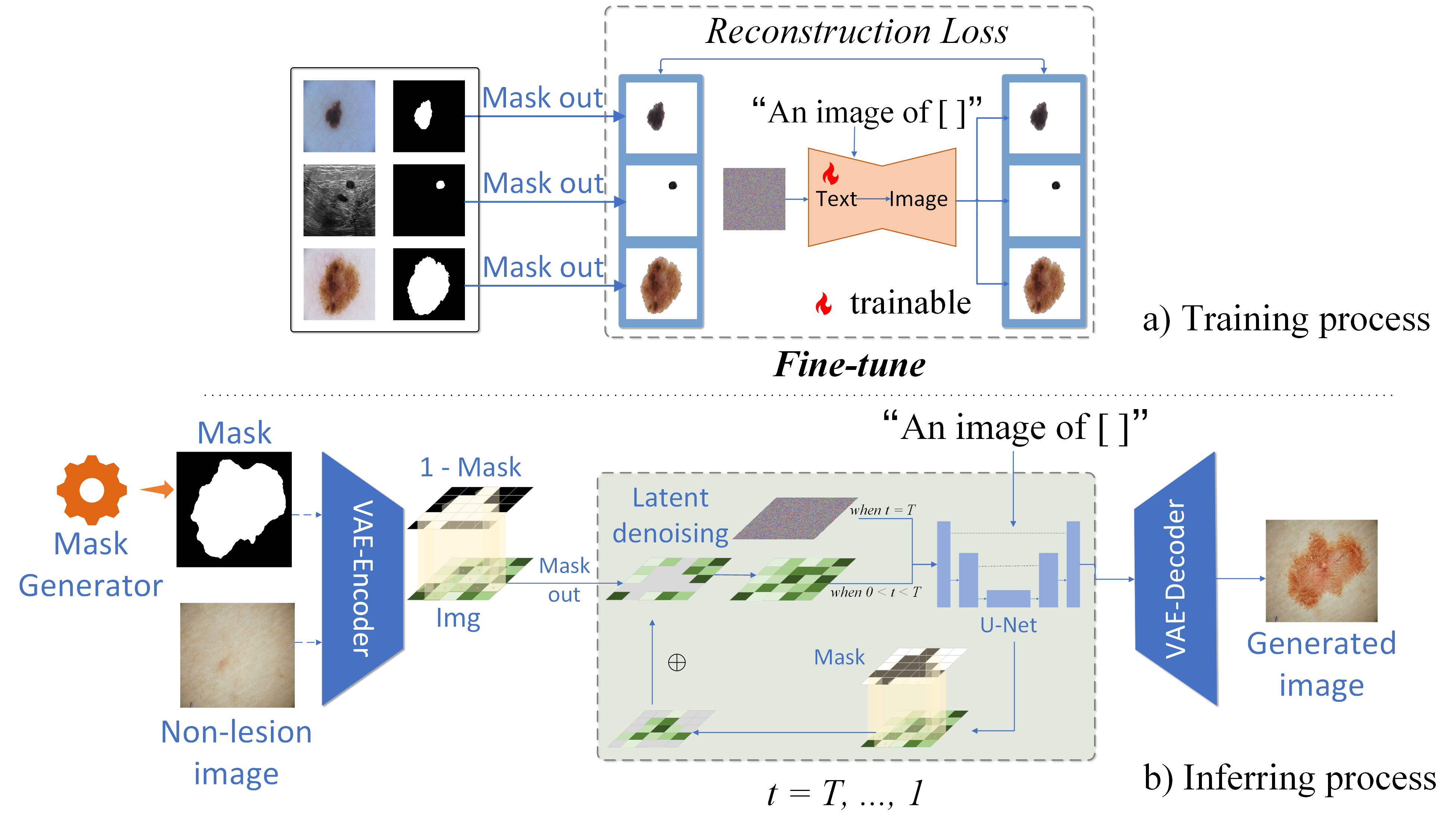} % Reduce the figure size so that it is slightly narrower than the column.
\caption{
Model structure. The figure is divided into parts a and b. Part a of the figure represents the training process, and part b represents the inference process. Both processes are effective, fast, and do not take much memory .}
\label{fig}
\end{figure*}
With the rise of Transformer-based architectures, Vision Transformer (ViT) \cite{vit} and its variants have been adapted for segmentation, yet their heavy data requirements and susceptibility to overfitting on small datasets remain challenges. Deep learning methods automatically learn feature representations without manual design, enabling robust performance in complex scenarios. However, medical images differ from natural images in data specificity, privacy constraints, and limited public availability, resulting in smaller datasets. This scarcity conflicts with the data-hungry nature of deep learning models, particularly Transformer-based architectures, making it difficult to achieve satisfactory segmentation with few labeled images.

To address data scarcity, diffusion models have emerged as powerful tools for synthetic data generation. Milestone works like DDPM \cite{ddpm} and DDIM \cite{ddim} established frameworks for iterative denoising, while Stable Diffusion improved efficiency via latent space mapping. Further, fine-tuning methods such as DreamBooth \cite{dreambooth} and LoRA \cite{lora} adapt pre-trained models to downstream tasks, and controllable generation techniques like ControlNet \cite{controlnet} enable structural guidance. However, these methods predominantly target natural images or classification tasks, with limited exploration of medical image-mask pair generation for segmentation. This limitation stems from fundamental differences between natural and medical imaging domains. 
% Natural images typically exhibit high intra-class variability but tolerate structural ambiguities, whereas medical images demand strict anatomical consistency and precise spatial correspondence between images and masks. Moreover, medical datasets face challenges such as stringent privacy regulations, labor-intensive expert annotations, and heterogeneous imaging modalities (e.g., MRI, CT, ultrasound), which hinder the direct adaptation of natural image generation paradigms. Existing methods often prioritize photorealism over anatomical fidelity or rely on class-level semantics unsuitable for pixel-wise segmentation tasks.

Current medical image generation methods \cite{dfclass,Roentgen,25,27} focus on classification or single-image synthesis, lacking mechanisms to produce paired images and masks. This gap hinders their utility for segmentation tasks. Additionally, fine-tuning diffusion models on limited medical data remains resource-intensive and underexplored. 
Some existing work \cite{26} can also generate the corresponding masks, though they rely on DDPM trained from scratch, which can be resource-intensive.

Generating synthetic medical image-mask pairs for segmentation introduces unique challenges: (1) Structural dependency: Synthetic images must preserve precise spatial relationships between organs or lesions and their corresponding masks, requiring pixel-level alignment beyond semantic plausibility. (2) Domain specificity: Medical imaging modalities exhibit distinct noise patterns and intensity distributions that synthetic data must replicate to avoid domain shifts. (3) Data efficiency: Fine-tuning large generative models on limited medical data risks overfitting or mode collapse, especially when training pairs number in the hundreds.
% (4) Clinical relevance — Generated anomalies (e.g., tumors) must follow pathological progression patterns rather than arbitrary appearances.

To bridge these gaps, we propose a lightweight, data-efficient method for controllable medical image-mask pair generation. Our method fine-tunes Stable Diffusion with limited data (under 30 minutes and 24GB memory) and uses automated quality assessment protocol filters to enhance reliability and diversity. In the inference phase, we use guide mask for controllable generation to achieve controllable shape and location of the lesion area. We also use a lightweight diffusion model as a mask generator to improve the versatility of the generated images. Experiments on five segmentation tasks demonstrate that models trained with our synthetic data achieve an average 3$\%$ accuracy improvement. 

Our contributions are:
1: A resource-efficient fine-tuning framework for medical image-mask pair generation.
2: A lightweight diffusion model with quality screening tailored for segmentation.
3: Empirical validation of synthetic data efficacy in downstream tasks.

\section{Method}
We present our work from two aspects: the paradigm of controlling diffusion models to generate reliable medical training data and the strategy of data selection. Our medical image generation framework with structural dependency and domain specificity requires only a few image-mask pairs. During training, a dynamic adaptive guiding mask highlights lesion regions, enabling fine-tuning of the Stable Diffusion model to focus on domain-specific feature learning. For inference, this adaptive mask guidance mechanism spatially constrains generation within predefined anatomical regions, ensuring precise controlled synthesis. We further develop a lightweight stochastic mask generator that produces both lesion-repairing patterns and anatomically plausible non-lesion maps, effectively expanding the data diversity. The post-processing phase incorporates an automated quality assessment protocol to ensure image fidelity, complemented by morphological corrosion operations to refine annotation boundaries.
\begin{figure*}[htb]
\centering
\includegraphics[width=0.9\textwidth,height=7.5cm]{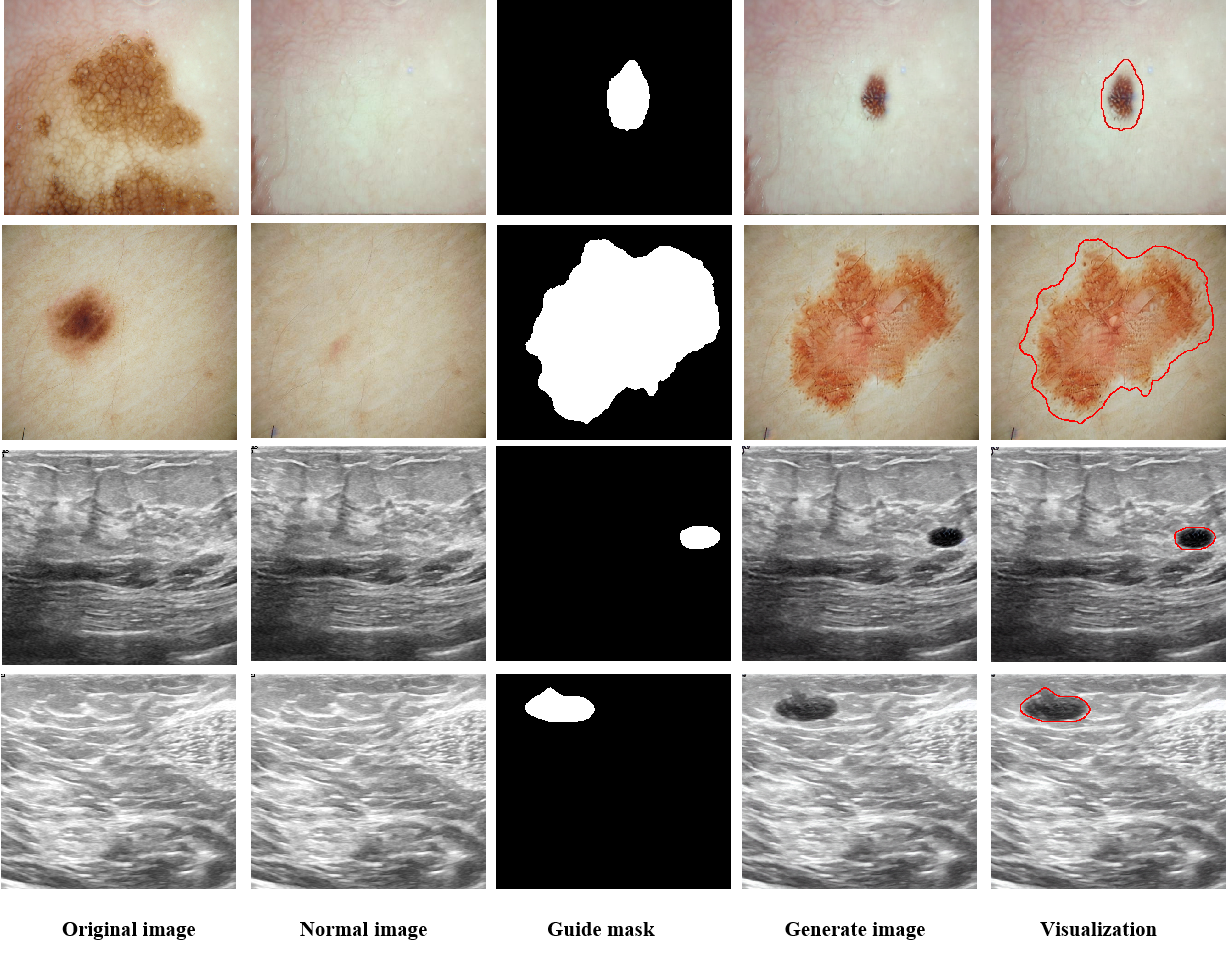} 
\caption{
The figure presents generated results from our method. The first two rows display skin images, where background images are restored from originals using our approach. The last two rows show breast ultrasound images; since background images are included in the dataset, these remain identical to the originals. Generated images are produced by applying guiding masks to background images, with the final column demonstrating controllable generation results.}
\label{fig1}
\end{figure*}
\subsection{Image Generation Framework}
\textbf{Controlled generation based on limited number of data
}Training a diffusion model from scratch with limited data (tens to thousands of samples) is challenging. Instead, we fine-tune the Stable Diffusion 1.5 \cite{sd1.5}. During training, we use specialized trigger words to reference new images, allowing the fine-tuned model to generate images corresponding to these concepts during inference, similar to the Textual Inversion \cite{textinversion} technique. However, unlike Textual Inversion, we unfreeze the parameters of U-Net for training during the training phase.

The controlled generation paradigm consists of training and inference processes. \textbf{In training}, the model focuses on lesion regions rather than the entire image. The original image and its corresponding mask are used to extract the lesion region, which is then fine-tuned. And the input to the model is noise, which generates images guided by specific prompt, and then the generated images are used to do a reconstructed loss calculation with the original images provided by us. After such training, it makes the fine-tuned model to generate a specific type of image under a specific prompt. A text prompt, such as "An image of hta," is designed to map to the lesion area, avoiding conflicts with Stable Diffusion's original vocabulary. This process allows the model to concentrate on the lesion region, requiring only around 30 samples and completing in 20 minutes with less than 24GB memory usage.
\begin{equation}
\mathcal{L}=\mathbb{E}_{t,X,c,\epsilon}[w_t||M \odot (\hat{X_\theta}(\alpha_tX + \beta_t \epsilon,c)-X)||_2^2]
\end{equation}

The equation 1 is the loss function for the training phase. $\hat{X_\theta}$ is a pre-trained text-to image diffusion model like Stable Diffusion 1.5, $c$ is a conditioning vector made by text prompt, $X$ is the ground-truth image, $\alpha_t,\beta_t,w_t$ are terms that control the noise schedule and sample quality, $M$ is the region mask matrix and $\epsilon$ is an initial noise map.

\textbf{For inference}, three components are needed: the fine-tuned model, a background (non-lesion) image, and a mask. The background image serves as the backdrop, while the mask directs the lesion region's location and shape, forming an image-mask pair for subsequent experiments. During inference, the model generates images in specified regions while maintaining the integrity of the rest of the image. The denoising process uses two latent vectors: one from the original image and one from the preceding denoising step. These vectors are combined to produce the intermediate denoised image, ensuring controlled generation. The full structure can be seen in Fig.1.
\begin{equation}
  x_{t-1} = M \odot[ \frac{1}{\sqrt{\alpha}_{t}}(x_{t}-\frac{\beta_{t}}{\sqrt{1-\bar\alpha_{t}}}\epsilon_{\theta}(x_{t},t))]+(1-M)\odot \mathcal{F}(x_{t}^{prev})   
\end{equation}

The denoising process in the inference phase is formulated as equation 2. $t$ is timesteps,  $M$ is region mask matrix, ${\alpha}_{t}, {\beta_{t}}$ are diffusion scheduling coefficients, $\epsilon_{\theta}$ is noise predicted by U-Net, $x_{t}^{prev}$ is latent variable state from the previous step and $\mathcal{F}$ is feature preservation function.

\textbf{Explanation of Model Components} Mask Generator: It is designed to produce highly diverse masks for guidance purposes. Implemented as a DDPM based on UNet architecture,trained on mask images. The network architecture consists of four hierarchical layers with progressively increasing channel dimensions [64,128,256,512]. Non-lesion Image Generator: Built upon Stable Diffusion 1.5 architecture,in the training phase, we use the data pairs: lesion Image, and invert the mask (swapping 0, 1 values). Then perform the original fine-tune, the inverted mask allows the model to learn how to generate healthy regions. In the inference phase, the input is the lesion image with original mask, like Fig.1 inference phase, the model can repair the lesion region to a healthy region in the mask region.

\textbf{Diversity of generation} In order to further enhance the diversity of generations, we adopt a dual-pronged approach to address the diversity problem.

% The first approach is to enhance the diversity of lesion-free background maps, which are normal images. In the event that the dataset contains lesion-free images, these lesion-free images can be employed directly. In the event that this is not the case, the diffusion model can be employed for the purpose of simple image restoration. However, when utilising a mask in the training process, the mask is flipped so that the diffusion model focuses on the normal region of the image. This enables the fine-tuned diffusion model to generate the normal region, which is then employed for inference purposes. The image in the training set with the mask input will be used by the model to generate an image of the normal region in the defective region. This will result in an image that repairs the defective or diseased region, that is, the background image without lesions. The effect of this process can be seen in Fig 3. Additionally, other techniques for image restoration are available.

% The second approach is to enhance the diversity of the mask. Although enhance the diversity by flipping, corrosion and other forms of image augmentation is feasible based on only the original training dataset of masks and images, it is considered that such diversity is still limited. Therefore, a small scale of diffusion model is introduced for mask data learning which can produce various masks. The generated masks is then used to guide the diffusion model to generate more medical images. In this way, the diversity of the generated images is greatly increased.

The first strategy enhances lesion-free background diversity through conditional image restoration. When datasets contain background images, these are directly utilized; otherwise, a diffusion model trained with inverted masks (focusing attention on healthy regions) reconstructs anatomically plausible tissue (Fig.2). This enables defect repair by generating non-lesion regions guided by input masks, ultimately producing pathology-free backgrounds.

The second strategy addresses mask diversity limitations of conventional augmentations (flipping/erosion) by training a compact diffusion model specifically for mask synthesis. These generated masks condition the main diffusion model, creating medically varied images that surpass traditional augmentation constraints, significantly expanding dataset diversity.
\begin{table*}[t]
\centering
\caption{Segmentation performance results. There are three experiments for each of these methods, original images, original images plus 1500 generated image pairs , and original images plus 2750 generated image pairs.The values in the table are Dice.}
\scriptsize
\begin{tabular}{|c|c|c|c|c|c|c|}
\hline
Method &    & PH$^2$& ISIC-2017 & ISIC-2018 & BUSI & DDTI   \\
 \hline
 & original  & 88.97&	83.05& 85.01&66.72&79.89  \\
 UPerNet  &original+1500 & 92.04	&		83.37	&	86.39	&	66.76	&	77.96  \\
 &original+2750 & 91.47	&		83.45	&	87.32	&	71.54	&	82.34 \\
 \hline
 & original  & 88.1	&		83.84	&	86.67	&	70.55	&	77.18  \\
 
 DeepLabV3 Plus  &original+1500 & 89.04	&		84.11	&	87.19	&	73.08 	&	78.84  \\
 &original+2750 & 89.53	&		84.7	&	87.87	&	74.05 	&	\textbf{83.01} \\
 \hline
 & original  & 89.46	&		81.19	&	86.03	&	59.55	&	74.03 \\
 
 Swin Transformer  &original+1500 & 91.38	&		81.57	&	86.68	&	62.38	&74.78	  \\
 &original+2750 & 92.16	&		84.28	&	86.55	&	63.63	&	75.22  \\
 \hline
 & original  & 94.26	&		83.77	&	87.43	&	77.01	&	79.45  \\
 
 nnU-Net &original +1500 & \textbf{94.75}	&		84.17	&	88.06	&	78.1	&	80.51  \\
 &original+2750 & 94.72	&		\textbf{84.92}	&	\textbf{88.15}	&	\textbf{78.69}	&	81.67  \\
\hline

\end{tabular}
%}

\label{table1}
\end{table*}

\subsection{Automated Quality Assessment Protocol}
Due to variability in generated data quality, a filtering process is implemented. Generated images should exhibit a degree of similarity with real images but not be too similar or dissimilar. The DINOv2 \cite{dinov2} model is used as a feature extractor to calculate cosine similarity between generated and original images. Images with excessively high or low similarity scores are excluded.

Additionally, a corrosion operation is applied to the mask edges to improve the fit between the mask and the generated lesion region, enhancing annotation quality. Ablation experiments in the experimental chapter demonstrate the validity of this filtering process.

This paradigm enables controlled generation with limited samples, applicable across diverse backgrounds and lesion types, locations, and shapes.

In the experimental chapter, some relevant ablation experiments are conducted to demonstrate the validity of data  filtering.
\begin{table*}[t]
\centering
\caption{Comparison of controllable generation methods. This includes the ControlNet and T2i-Adapter methods, and the results on the ISIC-2017 and BUSI datasets. Bolded numbers are the best values.The values in the table are Dice}
\scriptsize
\begin{tabular}{|c|c|c|c|c|c|c|}
 \hline
   Dataset &Method&  & UPerNet & DeepLabV3 Plus & Swin Transformer & nnU-Net \\
    \hline
  &   & Original  & 83.05 & 83.84 & 81.19 & 83.77 \\
  \cline{2-7}
  &  ControlNet & original+1500 & 83.29 & 83.31 & 81.06 & 83.67 \\
  &   & original+2750 & 83.4 & 83.77 & 81.74 & 84.48\\
    
ISIC-2017  &  T2I-adapter & original+1500 & \textbf{83.54} & 83.72 & 81.22 & 84.47 \\
  &   & original+2750 & 82.84 & 82.71  & 79.01 & 84.26\\
     
  &  Ours &original+1500 & 83.37  & 84.11   & 81.57 & 84.17\\
  &   & original+2750 & 83.45 & \textbf{84.7}  & \textbf{84.28} & \textbf{84.92}\\
     \hline
   &       & Original & 66.72 & 70.55 & 59.55 & 77.01 \\
     \cline{2-7}
  &  ControlNet & original+1500 & 69.3  & 71.76 & 59.69 &78.22 \\
   &  & original+2750 & 69.69 & 70.33 & 57.9 & 78.04\\
    
BUSI   & T2I-adapter & original+1500 & 69.42 &  72.41& 59.09 & 76.78 \\
   &  & original+2750 & 69.81 & 72.54  & 60.93& 78.30\\
   
  & Ours & original+1500 & 66.76 & 73.08  & 62.38 & 78.1\\
   &  & original+2750 & \textbf{71.54} & \textbf{74.05 }  & \textbf{63.63} & \textbf{78.69}\\
\hline
\end{tabular}
%}

\label{table2}
\end{table*}

\section{Experiments}
The experiment section is divided into three parts: datasets and baselines, results, and ablation experiments. These sections explore the impact of data filtering strategies, hyperparameters, and the use of generated data on the final segmentation results.
\subsection{Datasets and baselines}
Five publicly available datasets were used: ISIC-2017 \cite{isic2017}, ISIC-2018 \cite{isic2018}, PH2 \cite{ph2}, BUSI \cite{busi}, and DDTI \cite{ddti}. These datasets cover various organs (skin, thyroid, breast) and modalities (picture, ultrasound). Each dataset was divided into training, validation, and test sets. For ISIC-2017 and ISIC-2018, the provided test sets were used directly. For PH2, BUSI, and DDTI, 20$\%$ of the data was allocated to the test and validation sets, with the remainder used for training.

Four segmentation models were chosen as baselines: UPerNet \cite{upernet}, DeepLabV3 Plus \cite{deeplabv3}, Swin Transformer \cite{swintr}, and nnU-Net. These models represent both traditional and state-of-the-art approaches in medical image segmentation. Each model was trained on the training sets of the five datasets, with the best-performing weights on the validation set selected for testing.

The experimental setup was divided into two phases: segmentation and generation. The segmentation phase used Python 3.8.19, Pytorch 1.13.1+cu117, a batch size of 8, an input image size of 512x512, and 800 epochs. The loss function combined focal loss and dice loss, with dice vs. iou as the evaluation criterion. The generation phase used Python 3.10.14 and Torch 2.2.2, with a batch size of 2, 2000 iterations, and an input image size of 512x512.

\subsection{Results}
The original line experiments used 30 image-mask pairs for fine-tuning. Subsequently, 50 images were generated based on 50 background images, with two sets of generation quantities: 1,500 and 2,750 images. These generated image-mask pairs were added to the training sets, while the validation and test sets remained unchanged.

As shown in Table 1, incorporating 1,500 and 2,750 additional images into the training sets resulted in higher DICE scores compared to the original line results. The improvement was more pronounced with more generated data pairs, as the increased diversity and quality of the training data enhanced the model's generalization and stability.
\begin{table*}[t]
\centering
\caption{The result of post-processing. Each method includes two comparisons: data filtering and corrosion mask. The baseline is the original image and 2750 generated images are added for the training. The evaluation metric utilized is the DICE score.}
\scriptsize
\begin{tabular}{|c|c|c|c|c|c|c|}
\hline
Method &    & PH2& ISIC-2017 & ISIC-2018 & BUSI & DDTI   \\
 \hline
 & Baseline  & 91.47&	83.45& 87.32 &71.54&82.34 \\
 UPerNet  &Baseline+filter &   93.76	&	84.11 	&87.88		&	 71.75 	&83.26	  \\
 &Baseline+corrode &  93.23	&	84.57		&	 87.32 	&	 71.76	&	 83.46   \\
 \hline
 & Baseline  & 89.53	&		84.7 	&	87.87	&	74.05  	&	83.01  \\
 DeepLabV3 Plus  &Baseline+filter &90.91	&	84.74		&	87.9	&	74.54	&	 83.25   \\
 &Baseline+corrode & 91.09 	&	84.73		&	87.55	&	73.73	&	83.27\\
 \hline
 & Baseline  & 92.16	&		84.28 	&	86.55 	&	63.63	&	75.22  \\
 
 Swin Transformer  &Baseline+filter & 92.1	&84.33			&86.33 		&	64.03	&75.26	 \\
 &Baseline+corrode &91.19 &		84.3	&	86.74 	&	 63.51	&75.37	 \\
 \hline
 & Baseline  & 94.72	&		84.92	&	88.15	&	78.69	&	81.67 \\
 nnU-Net &Baseline+filter & 94.69	&	84.96		&88.24		&	78.88	&81.32	  \\
 &Baseline+corrode&95.02 	&	84.70		&	88.87	&	78.34	&	82.25  \\
\hline

\end{tabular}
%}

\label{table3}
\end{table*}
\begin{figure}[htb]
\centering
\includegraphics[width=0.85\columnwidth,height=3.5cm]{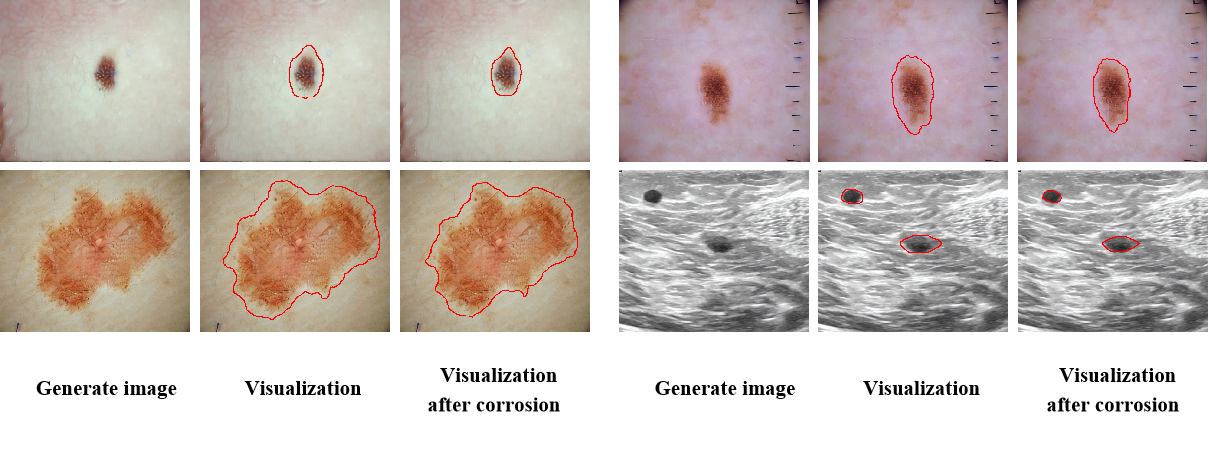} % Reduce the figure size so that it is slightly narrower than the column. Don't use precise values for figure width.This setup will avoid overfull boxes.
\caption{Erosion effect. From left to right: generate image, visualization of image and corresponding mask, and visualization after erosion mask.}
\label{fig2}
\end{figure}
Two controllable generation methods, ControlNet and T2I-Adapter \cite{t2i}, were also tested using the same number of generated image-mask pairs. These methods provided some assistance for downstream segmentation tasks but were less stable and produced suboptimal results compared to our approach. This discrepancy may be due to the limited training data (30 pieces), which was insufficient for ControlNet and T2I-Adapter to achieve optimal results.

% \begin{table}[t]
% \centering
% \caption{Results of ablation experiment. Among them, 20, 30, and 50 are the normal number of images used during generation. Each experiment is generated by adding 2750 images to the original images.The values in the table are Dice.}
% %\resizebox{.95\columnwidth}{!}{
% \begin{tabular}{|c|c|c|c|c|}
% \hline
%     Dataset  &  Method   &20 & 30 &50 \\
% \hline   
%    & UPerNet    & 83.12 & 83.42  &\textbf{83.45}  \\
%  ISIC-2017  & DeepLabV3 Plus  &  84.09 & 84.58  &\textbf{ 84.7} \\
%     &Swin Transformer   & 81.6  & 81.78 &\textbf{84.28} \\
%     &nnU-Net   & 85.31 & \textbf{85.37} & 84.92 \\
% \hline    
%    & UPerNet    & 86.25  &  86.7 &\textbf{88.45} \\
% ISIC-2018   & DeepLabV3 Plus   &  86.55 &  86.71  & \textbf{87.87} \\
%    & Swin Transformer   &  86.18 & \textbf{86.58} & 86.55 \\
%    & nnU-Net   & 87.63 & 87.89 & \textbf{88.15} \\
% \hline
% \end{tabular}

% \label{table4}
% \end{table}
\begin{table}[t]
\centering
\caption{Results of ablation experiment. Among them, 20, 30, and 50 are the number of background images used during generation. Each experiment is adding 2750 generatio images to the original images.The values in the table are Dice.}
\scriptsize
%\resizebox{.95\columnwidth}{!}{
\begin{tabular}{|c|c|c|c|c||c|c|c|c|}
\hline
    Dataset  &  Method   &20 & 30 &50& Method   &20 & 30 &50\\
\hline   
   & UPerNet    & 83.12 & 83.42  &\textbf{83.45} &Swin Transformer   & 81.6  & 81.78 &\textbf{84.28} \\
 ISIC-2017  & DeepLabV3 Plus  &  84.09 & 84.58  &\textbf{ 84.7}&nnU-Net   & 85.31 & \textbf{85.37} & 84.92 \\
    % &Swin Transformer   & 81.6  & 81.78 &\textbf{84.28} \\
    % &nnU-Net   & 85.31 & \textbf{85.37} & 84.92 \\
\hline    
   & UPerNet    & 86.25  &  86.7 &\textbf{88.45} & Swin Transformer   &  86.18 & \textbf{86.58} & 86.55\\
ISIC-2018   & DeepLabV3 Plus   &  86.55 &  86.71  & \textbf{87.87}& nnU-Net   & 87.63 & 87.89 & \textbf{88.15}  \\
   % & Swin Transformer   &  86.18 & \textbf{86.58} & 86.55 \\
   % & nnU-Net   & 87.63 & 87.89 & \textbf{88.15} \\
\hline
\end{tabular}

\label{table4}
\end{table}
\subsection{Ablation experiments}
The ablation experiments focused on two aspects: the number of background images used for generation and the role of data filtering.

For the number of background images, experiments were conducted with 20, 30, and 50 background images. As shown in Table 4, the segmentation results improved with more background images, as increased diversity in the generated data led to better model performance.

Data filtering was applied to the generated images, removing those that were too similar or dissimilar. As shown in Table 3, filtering reduced the amount of generated data but improved segmentation performance by enhancing the quality of the training dataset. Additionally, corroding the masks of the generated images further improved the results, as it made the masks more accurate and better aligned with the lesions in the images.

In summary, the experiments demonstrated that increasing the diversity and quality of training data through controlled generation and filtering significantly improved segmentation performance. Our approach outperformed other controllable generation methods, highlighting its potential for addressing the challenge of limited medical imaging data.

 % Meanwhile, to verify the improvement of segmentation that the generated data can bring compared to the real data, we set up five training sets with different number of real images, which are 30 images, 60 images, 100 images, 200 images and the whole number of original dataset. And also, 30 real images plus generated data are used to form a combined training set for comparison to check how the generated data with real data improves the segmentation model effect compared to only real data. The results are shown in Table x.

\section{Conclusion}
This paper proposes a resource-efficient controllable generation paradigm for diffusion models using limited data. By fine-tuning Stable Diffusion with minimal training data, our method produces high-quality image-mask pairs validated through downstream segmentation tasks. The framework demonstrates hardware-friendly implementation with low-resource training/inference requirements. Extensive validation across five real-world datasets using four segmentation architectures confirms the effectiveness of the synthetic data.

\noindent
\textbf{Acknowledgments} 
This work is supported by 2023 Shenzhen sustainable supporting funds for colleges and universities (20231121165240001), Shenzhen Science and Technology Program (JCYJ20230807120800001), Guangdong
Provincial Key Laboratory of Ultra High Definition Immersive Media Technology 
(2024B
1212010006).

\end{document}